\documentclass[journal]{IEEEtran}

\input{config.sty}
\input{glossary.sty}

\title{Automated Constraint Specification for Job Scheduling by Regulating Generative Model with Domain-Specific Representation}

\author{
Yu-Zhe Shi, Qiao Xu, Yanjia Li, Mingchen Liu, Huamin Qu,\\
Lecheng Ruan and Qining Wang,~\IEEEmembership{Senior Member, IEEE}%
\thanks{This work is partially supported by the National Natural Science Foundation of China under Grants 52475001 and RGC GRF Grant 16210321. Yu-Zhe Shi and Qiao Xu contributed equally. (Corresponding author: Lecheng Ruan)}
\thanks{Yu-Zhe Shi, Qiao Xu, Lecheng Ruan, and Qining Wang are with School of Advanced Manufacturing and Robotics, Peking University, Beijing 100871, China.}
\thanks{Yu-Zhe Shi, Yanjia Li, and Huamin Qu are with the Department of Computer Science and Engineering, School of Engineering, The Hong Kong University of Science and Technology, Hong Kong SAR.}
\thanks{Mingchen Liu is with the School of Computer Science and Technology, Huazhong University of Science and Technology, Wuhan 430074, China.}
}

\begin{document}

\maketitle

\begin{abstract}
Advanced Planning and Scheduling (APS) systems have become indispensable for modern manufacturing operations, enabling optimized resource allocation and production efficiency in increasingly complex and dynamic environments. While algorithms for solving abstracted scheduling problems have been extensively investigated, the critical prerequisite of specifying manufacturing requirements into formal constraints remains manual and labor-intensive. Although recent advances of generative models, particularly Large Language Models (LLMs), show promise in automating constraint specification from heterogeneous raw manufacturing data, their direct application faces challenges due to natural language ambiguity, non-deterministic outputs, and limited domain-specific knowledge. This paper presents a constraint-centric architecture that regulates LLMs to perform reliable automated constraint specification for production scheduling. The architecture defines a hierarchical structural space organized across three levels, implemented through domain-specific representation to ensure precision and reliability while maintaining flexibility. Furthermore, an automated production scenario adaptation algorithm is designed and deployed to efficiently customize the architecture for specific manufacturing configurations. Experimental results demonstrate that the proposed approach successfully balances the generative capabilities of LLMs with the reliability requirements of manufacturing systems, significantly outperforming pure LLM-based approaches in constraint specification tasks.      
\newline

\emph{Note to Practitioners}---This paper presents a practical solution for automating the conversion of raw manufacturing information into job scheduling specifications, addressing a common challenge in implementing APS systems. The proposed architecture can process diverse manufacturing documentation formats, from semi-structured route sheets to natural language instructions, while ensuring reliability through domain-specific representations. Manufacturing practitioners can use this system to reduce the manual effort in specifying digitalized constraints for production, particularly beneficial for facilities with frequent requirement changes or small-batch, multi-variety production. The system's ability to automatically adapt to different manufacturing scenarios makes it accessible without requiring extensive programming expertise, offering a practical balance between automation and accuracy in production planning.
\end{abstract}

\begin{IEEEkeywords}
Smart Manufacturing; Constraint Specification; Domain-Specific Representation
\end{IEEEkeywords}

\IEEEpeerreviewmaketitle

\section{Introduction}

Smart manufacturing has become a cornerstone of industrial development, promising enhanced productivity and adaptability through digital transformation~\cite{kusiak2018smart}. This manufacturing paradigm represents a fundamental shift from traditional production systems, incorporating advanced technologies to create more intelligent and responsive operations~\cite{conti2022human, balta2023digital, chen2022resource}. As manufacturers face increasing pressure to handle product variety and shorter delivery times, the conventional rigid production models are giving way to \ac{fms}~\cite{browne1984classification}. These systems must efficiently process multiple jobs with different specifications, processing requirements, and priorities---all while sharing limited manufacturing resources. Accordingly, \ac{aps} system~\cite{hvolby2010technical} emerges as a crucial decision-making process in smart manufacturing, as it determines how effectively manufacturing resources are utilized to meet production objectives.

The implementation of an \ac{aps} system comprehends two major phases: (i) the \emph{specification} of real-world manufacturing constraints, such as procedural dependency for production and availability of factory resource, into a mathematical \ac{jsp} formulation~\cite{dauzere2024flexible, xiong2022survey}; and (ii) the \emph{solution} for the formulated \ac{jsp}~\cite{gao2019review, cunha2020deep}. In the previous research paradigm, manufacturing constraints are typically abstracted and standardized to emphasize the scheduling fundamentals while deliberately excluding production-scenario-specific context such as material characteristics, machine specifications, and product requirements~\cite{yao2023novel, fontes2024energy}. Under this setting, diverse \ac{jsp} solvers have been successfully developed for specific problem characteristics and instances~\cite{fontes2023hybrid, zhang2023survey}.

\begin{figure*}[ht]
    \centering
    \includegraphics[width=\linewidth]{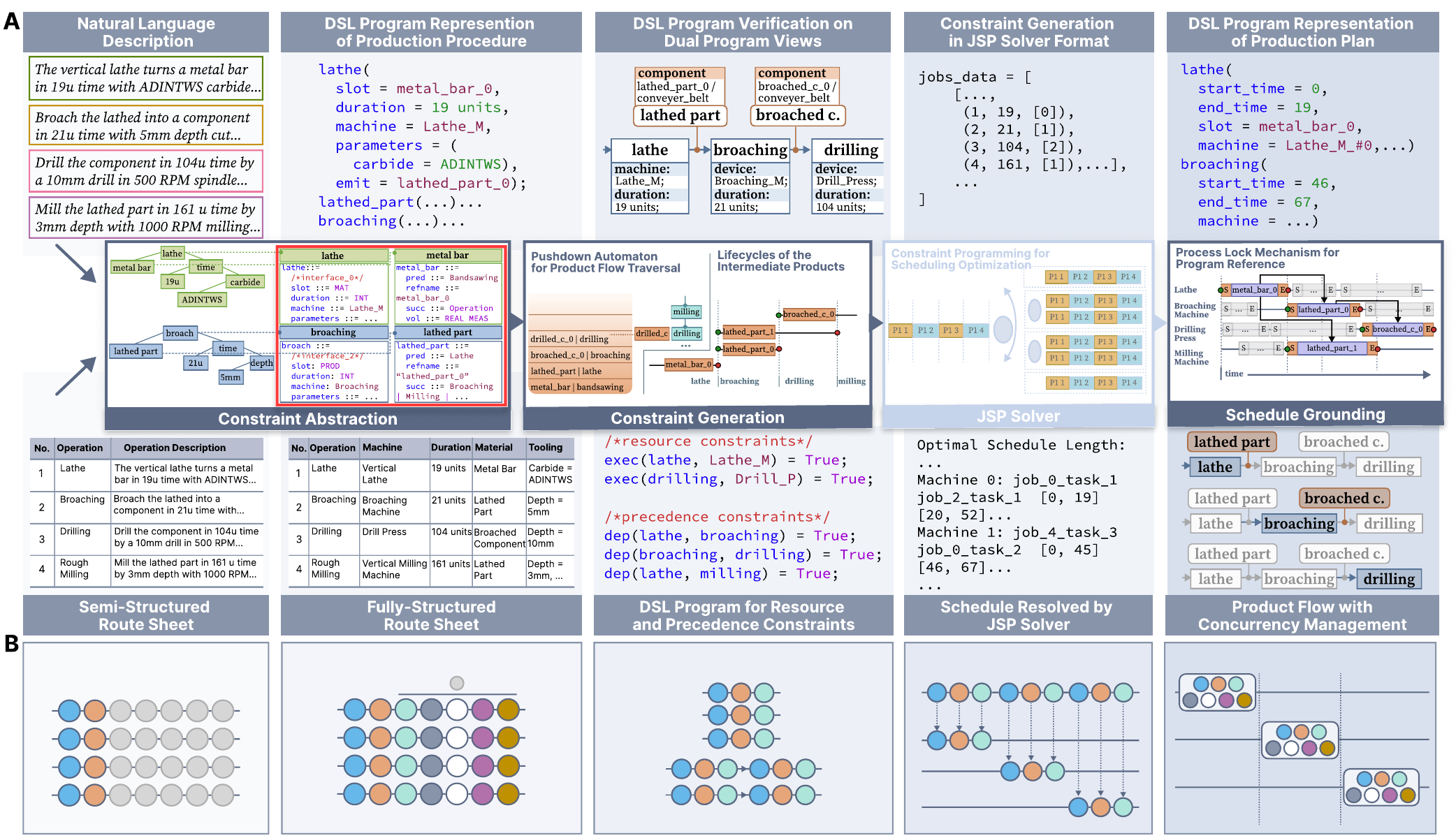}
    \caption{\textbf{Illustration of the proposed constraint-centric architecture.} \textbf{(A)} This panel presents a running example that illustrates the complete information flow, transitioning from two formats of procedures to a grounded production plan (depicted in the top and bottom rows). It also outlines the three modules involved: \textbf{\hmodule{constraint abstraction}}, \textbf{\hmodule{constraint generation}}, and \textbf{\hmodule{schedule grounding}}, along with their corresponding core working mechanisms (shown in the middle row). The \acs{dsl} program specified for the manufacturing scenario is \textbf{\hdslbox{highlighted}} as the primary driving force of the architecture. \textbf{(B)} This panel provides an intuitive visualization of the procedural transmission of information throughout the architecture's process. Progressing from left to right, the sequence of states includes the original procedures, fully-structured route sheet, generated constraint, \ac{jsp}-solver-generated schedule, and ultimately, the production plan. The information is color-coded according to its type, such as \textbf{\hoperation{operation name}}, \textbf{\hmachine{machine}}, \textbf{\hduration{duration}}, and \textbf{\hunspecified{unspecified information}}, to enhance understanding.}
    \label{fig:pipeline}
\end{figure*}

However, abstracted constraints for a \ac{jsp} solver do not emerge spontaneously from real-world manufacturing scenarios; they require meticulous development and formalization by experienced manufacturing experts~\cite{li2024flexible}. This manual specification process was historically manageable in traditional manufacturing settings with stable production scopes, as it was essentially a one-time effort with long-term utility. However, the emergence of smart manufacturing has fundamentally transformed this landscape through several challenges. First, the shift toward small-batch, multi-variety production demands extensive constraint specifications to accommodate diverse processing requirements~\cite{tian2023dynamic}. Second, the inherently dynamic nature of modern production necessitates constant updates of constraints~\cite{zhao2023drl}. Third, manufacturing knowledge exists in heterogeneous formats across different sources, ranging from semi-structured route sheets to \ac{nl} instructions~\cite{wang2022dynamic, zhang2019semantics}, creating significant barriers to efficient constraint extraction. While researchers have made considerable progress improving \ac{jsp} solver efficiency, the manual constraint specification process has emerged as the critical bottleneck in \ac{aps} implementation. Consequently, there is a growing need to automate the constraint specification process, and furthermore the entire workflow---from abstracting heterogeneous constraint rules to generating solver specifications and ultimately grounding the resulting schedules back to interpretable production plans.

With the rapid advancement of \ac{genai} techniques, leveraging \acp{llm} to automate constraint specification based on heterogenous manufacturing data and knowledge appears promising. Recent studies have demonstrated \acp{llm}' capability to solve abstract optimization problems from mathematical word problem-style \ac{nl} descriptions~\cite{ramamonjison2022augmenting,ramamonjison2023nl4opt}, suggesting their potential applicability for specifying the constraints~\cite{ahmaditeshnizi2024optimus,xiao2023chain,huang2025orlm,jiang2025llmopt,zhang2025decision}. However, despite \acp{llm}' strengths in information extraction and \ac{nl} understanding, their direct application to manufacturing constraint specification faces several fundamental challenges: (i) the intrinsic ambiguity of \ac{nl}-based raw manufacturing data such as route sheets~\cite{russell1923vagueness}; (ii) the non-deterministic nature of generative models like \acp{llm}~\cite{ouyang2024empirical}; and (iii) the lack of fine-grained domain-specific knowledge regarding the specific product categories and factories in the training data of \acp{llm}~\cite{cannas2024artificial}. While such characteristics might be acceptable or even beneficial in creative applications like artwork composition, manufacturing systems demand absolute precision, reliability, and detailed domain expertise~\cite{shi2023complexity}. Specifically, generated constraints must strictly align with factory resources and product-specified production procedure, while production plans must precisely define execution configurations. This fundamental mismatch between manufacturing's reliability requirements on constraints and the capabilities of pure \ac{llm}-based solutions suggests the necessity of an external architecture atop the \ac{llm}-based workflow to regulate the \ac{llm} performance in the manufacturing scenarios.

To address these challenges, this paper proposes a \emph{constraint-centric architecture} that regulates the \ac{llm} to perform automated reliable constraint specification from raw manufacturing knowledge and data. The architecture defines a hierarchical structural space that systematically organizes manufacturing constraints across three distinct levels~\cite{shi2024abstract}. The top level captures global operation dependencies and resource relationships. The middle level handles the context-specific execution configurations, while the bottom level manages detailed scheduling parameters and production specifications. This hierarchical space is gradually constructed upon a domain-specific representation, implemented through \acp{dsl}~\cite{fowler2010domain} for their properties that naturally align with manufacturing constraints~\cite{tarski1946introduction, russell2010artificial}---(i) the compact domain-specific features to avoid redundancy of constraint representation~\cite{karsai2009design}; (ii) the support of flexible, syntactical combination of constraints~\cite{chomsky1957syntactic}; and (iii) the maintenance of reliability by structural constraint representation~\cite{hopcroft1996introduction}---serving as an appropriate supplementary of \acp{llm}~\cite{shi2024expert}. To further adapt the proposed architecture across different manufacturing scenarios, such as varied factory configurations and diverse production procedures~\cite{mernik2005and}, and avoid the labor-intensive, case-specific, and costly \ac{dsl} crafting~\cite{shi2024autodsl,shi2025hierarchically,shi2025targeted}, we present an automated algorithm to efficiently adapt the architecture according to specific manufacturing configurations. Comprehensive experiments are conducted to verify our advantages over pure \ac{llm}-based approaches.

In this work, we address the automatic, entire-workflow constraint specification for production scheduling, based on a domain-specific representation to regulate \ac{llm} outputs, to achieve a balance between the generative power of \ac{genai} and the guardrail of reliability. Our contributions are three-fold: (i) we introduce and formulate the constraint-centric architecture for reliable constraint specification (\cref{sec:arch}); (ii) we develop an automatic adapter for customizing the proposed architecture across various production scenarios (\cref{sec:auto}); (iii) we integrate the architecture into the \ac{genai} pipeline for constraint specification, evaluate the pipeline on diverse manufacturing scenarios (\cref{sec:exp}), and demonstrate the usability and scalability of the proposed architecture (\cref{sec:res}).

\section{The Constraint-Centric Architecture}\label{sec:arch}

In this section, we introduce the constraint-centric architecture for entire-life-cycle constraint specification (see \cref{fig:pipeline}). Starting from an overview of the architecture (\cref{subsec:arch-overview}), we describe the three modules, including abstraction from the route sheets (\cref{subsec:arch-o2c}), specification for \ac{jsp} (\cref{subsec:arch-c2j}), and instantiation for interpretable production plan (\cref{subsec:arch-j2p}).

\subsection{Architecture Overview}\label{subsec:arch-overview}

The primary utility of the architecture lies in its ability to take new-coming production procedures as input and generate a corresponding scheduled production plan as output, tailored to the specific context of a concrete factory within a manufacturing domain. The overall input to the architecture can be classified into two types: (i) \ac{nl}-based production documents, which specify the target production procedure step-by-step in textual form; or (ii) semi-structured manufacturing route sheets, which contain extracted columns detailing the operation name and \ac{nl}-based operation descriptions. According to the \ac{sop} in manufacturing, production procedure descriptions must be transformed into fully-structured manufacturing route sheets, which serve as the foundation for scheduling. These route sheets typically include columns detailing machine names and operation durations. Therefore, to accurately capture such information within the production procedures, the first module compiles the unstructured or semi-structured descriptions into programs of the corresponding \ac{dsl} of the specific production scenario.

Subsequently, the second module works on the fine-grained route sheet and specifies the constraints for \ac{jsp}. This is implemented as verification over the \ac{dsl} programs, generating resource constraint programs and precedence constraint programs. These programs are then transformed into a format compatible with off-the-shelf \ac{jsp} solvers. 

Finally, the third module grounds the resulting schedule, output by the \ac{jsp} solvers, into production plans that are ready for further interpretation and execution by the \ac{cnc} systems within the factory environment. This step completes the concrete semantics of the operations and their corresponding execution configurations, transforming the semanticless schedule produced by the \ac{jsp} solvers. This is achieved by referencing the \ac{dsl} programs.

\subsection{Constraint Abstraction from the Production Procedures}\label{subsec:arch-o2c}

The first module takes the production procedures as input and outputs a fully structured representation of the target production procedures, specifically the complete manufacturing route sheet. This desired route sheet must accurately capture the operation name, required machine, standard duration, and execution configurations for each step. Given the intrinsic ambiguity of \ac{nl}~\cite{russell1923vagueness}, achieving this objective necessitates the precise parsing of \ac{nl}-based descriptions and the fine-grained representation of procedural knowledge. In precision-demanding scenarios like manufacturing, any deviation from the provided procedures or the factory’s conditions is inadmissible. To ensure both preciseness and usability, we opt to utilize domain-specific representation for production procedures, implemented through \acp{dsl}.

The working mechanism of this module can be formally expressed as $\text{CAP}=(\Upsilon \mid \Gamma, \mathcal{L})$, where $\Gamma=\{\gamma_1,\gamma_2,\dots,\gamma_{|\Gamma|}\}$ represents the given \emph{set of production procedures}, which outline the steps for producing specific products $\gamma_k$; $\mathcal{L}$ denotes the \ac{dsl} tailored to the manufacturing scenario and the factory. The outcome $\Upsilon=\{\mathcal{J},\mathcal{M}\}$ represents the set of fully structured manufacturing route sheets, where $\mathcal{J}=\{J_1,J_2,\dots,J_{|\mathcal{J}|}\}$ indicates the production procedures represented as \ac{dsl} programs, and $\mathcal{M}=\{M_1,M_2,\dots,M_{|\mathcal{M}|}\}$ denotes the \emph{factory}, equipped with a concrete set of machines.

The corresponding \ac{dsl} $\mathcal{L}=\{\mathcal{L}_o,\mathcal{L}_p\}$ is indicated by a pair of dual program views: the operation-centric program view $\mathcal{L}_o$ and the product-flow-centric program view $\mathcal{L}_p$. This design choice arises from the understanding that both operations and product flows are critical elements in production procedures, yet they cannot be tracked simultaneously because they are intricately intertwined with each other. When focusing on operations, the context becomes the input and output products; conversely, when focusing on products, the context shifts to the operations that yield and consume them, respectively. By adopting a dual representation of these two views, we are able to track the detailed execution of operations and the detailed transition of product flows in parallel, resulting in a precise and reliable model of the production procedure.

The operation-centric program view focuses on the execution context of an operation. It captures the availability of necessary materials as the precondition, the required machines, and their corresponding execution configurations as the program body, along with the expected output as the postcondition. This view serves as an \emph{interface} between the semantic identifier and the grounded instances of the operations, interpreting the purpose of the operation indicated by the procedure in terms of a specific execution context. For example, a milling operation can be performed on either a vertical or a horizontal milling machine, depending on the scale of the input material and the dimensions of the product. 

In contrast, the product-flow-centric program view emphasizes the states of the product as it progresses through the various operations, effectively modeling the product flow. Each flow unit is produced by a predecessor operation and consumed by a successor operation. A critical property of the product flow is its spatial-temporal continuity, whereby the transitions between the states of the product are largely driven by specific operations. This view acts as a \emph{pipe}~\cite{gabbrielli2010programming}, passing products along the temporal dimension and tracking the invariance of the entire procedure. For instance, to produce a part incorporating two different types of metal materials, the raw material input to the production line must consist of the two required materials; these materials cannot appear from nowhere but must be passed down the product flows.

The operation-centric program view, denoted as $\mathcal{L}_o=\{\mathcal{S}_o, \Lambda_o\}$, is characterized by the syntactic language feature set $\mathcal{S}_o$ and the semantic language feature set $\lambda_o$. The syntax $\mathcal{S}_o=(\varphi,\phi, \varphi^{\text{prec}},\varphi^{\text{post}}, \varphi^{\text{exec}})$ defines a structural space for encapsulating the precondition $\varphi^{\text{prec}}$, postcondition $\varphi^{\text{post}}$, and execution $\varphi^{\text{exec}}$ within the interface structure of the operation. Utilizing this syntax, an operation with the semantic identifier $\varphi$ is referenced through an interface $\phi$ to a set of execution contexts, represented as
\begin{equation}
    \langle\varphi\mapsto\phi\mapsto\{(\varphi^{\text{prec}},\varphi^{\text{post}}, \varphi^{\text{exec}})\}\rangle.
\end{equation}
The operation $\varphi$ can be grounded to a corresponding instance in any compatible execution context, echoing the idea of \emph{modular design}~\cite{abelson1996structure}. The semantics $\Lambda_o=(\Phi,\Phi^{\text{prec}},\Phi^{\text{post}},\Phi^{\text{exec}})$ specifies the permissible assignments of the fields and their corresponding values within the structural space defined by $\mathcal{S}_o$, where $\varphi\in\Phi$, $\varphi^{\text{prec}}\in\Phi^{\text{prec}}$, $\varphi^{\text{post}}\in\Phi^{\text{post}}$, and $\varphi^{\text{exec}}\in\Phi^{\text{exec}}$. Refer to \cref{sec:auto} for the automated design of such \acp{dsl}.

The product-flow-centric program view, denoted as $\mathcal{L}_p=\{\mathcal{S}_p, \Lambda_p\}$, is characterized by the syntactic language feature set $\mathcal{S}_p$ and the semantic language feature set $\lambda_p$. The syntax $\mathcal{S}_p=(\omega,\omega^{\text{pred}},\omega^{\text{succ}},\omega^{\text{prop}},\psi\langle\omega^{\text{pred}},\omega^{\text{succ}}\rangle)$ defines a structural space for encapsulating a selected set of key properties of the product $\omega^{\text{prop}}$, the predecessor $\omega^{\text{pred}}$, and the successor $\omega^{\text{succ}}$ within the interface structure of the product, which is inherited from the operation-centric program view. Using this syntax, a product with the semantic identifier $\omega$ is represented as $\langle\omega\mapsto(\omega^{\text{pred}},\omega^{\text{succ}},\omega^{\text{prop}})\rangle$. Additionally, there is a special syntactic feature $\psi\langle\omega^{\text{pred}}_t,\omega^{\text{succ}}_t\rangle$ that captures the pipe structure $\psi\langle\cdot,\cdot\rangle$ of the product flow at the unit level $\omega_t$, indicating potential $N$-predecessors-to-$M$-successors relationships within the product flow. If the product is transferred directly from the predecessor to the successor, the pipe forms a linear structure. In cases where two products are produced by two different predecessors and consumed by one successor, the pipe forms a \emph{``Y-shaped''} structure. This syntax can express any \emph{``$N$-to-$M$-intersection-shaped''} relationships along the product flow, \ie, $N$ products produced by $N$ different predecessors and consumed by $M$ different successors. Consequently, the pipe structure syntax provides a guardrail for accurately modeling product flows of varying complexity. The semantics $\Lambda_p=(\Omega,\Omega^{\text{pred}},\Omega^{\text{succ}},\Omega^{\text{prop}},\Psi)$ specifies the permissible assignments of the fields and their corresponding values within the structural space defined by $\mathcal{S}_p$, where $\omega\in\Omega$, $\omega^{\text{pred}}\in\Omega^{\text{pred}}$, $\omega^{\text{succ}}\in\Omega^{\text{succ}}$, $\omega^{\text{prop}}\in\Omega^{\text{prop}}$, and $\psi\in\Psi$. Please refer to \cref{sec:auto} for the automated design of such \acp{dsl}.

We implement the translation from the original procedures $\Gamma$ to the fully structured route sheets $\Upsilon$ inspired by the practice of Shi~\etal~\cite{shi2024expert}, to guarantee the completeness of syntax, the correctness of semantics, and the reliance of execution. Given an original description of production procedure $\gamma_k\in\Gamma$ for translation, we first parse the \ac{nl} sentences by an off-the-shelf tool and extract the actions accordingly~\cite{honnibal2015improved}. Then, the extracted actions are matched with the operation set $\Phi$ of the \ac{dsl}, according to both exact match score and semantic similarity. Afterwards, we extract the arrays of entities related to the extracted action $\mathcal{E}$ by an off-the-shelf \ac{llm}-based tool~\cite{xie2024self}, where we regard the output labels to the entities and relations as \emph{pseudo-labels} because they can possibly be noisy. On this basis, we can formulate the objective of this \ac{dsl} program synthesis tasks as
\begin{equation}
    \begin{aligned}
        &\ \arg\min_{^*\mathcal{L}(\Gamma),\mathcal{E}} \;D(\Gamma\parallel\Upsilon)\\
        s.t.& \quad\mathcal{L}=\{\mathcal{S}_o,\mathcal{S}_p,\Lambda_o,\Lambda_p\},
    \end{aligned}
\end{equation}
where $^*\mathcal{L}(\Gamma)=\{\mathcal{L}(\Gamma)\mid \Gamma\Rightarrow^*\mathcal{S}_o, \Gamma\Rightarrow^*\mathcal{S}_p, \mathcal{L}(\Gamma)\in\Lambda_o\cup\Lambda_p\}$ denotes the set of all possible \ac{dsl} program patterns generated by the procedures described by $\Gamma$. The divergence function $D(\cdot\parallel\cdot)$ possesses three indicators: (i) the selected program patterns should be as close as possible to the text span; (ii) the selected program pattern should be as similar as possible with the extracted subject-verb-object structure parsed from \ac{nl} description; and (iii) as many pseudo-labeled entities as possible should be mapped to the semantics space. 

\subsection{Constraint Generation for JSP Formulation}\label{subsec:arch-c2j}

The second module processes fully structured route sheets, derived from the original set of production procedures, and produces a specified set of constraints that formulates the \ac{jsp}. This specification is subsequently converted into a format compatible with off-the-shelf \ac{jsp} solvers, which are then utilized to determine the optimal schedule for the production procedures. The constraint generation mainly necessitates two types of constraints: resource constraints and precedence constraints. Ensuring the accuracy of the constraint generation is paramount to guaranteeing that the schedules generated by the \ac{jsp} solver are both meaningful and correct. This specification is achieved through contextualized \ac{dsl} program verification, based on the \ac{dsl} programs representing the route sheet.

The working mechanism of this module can be formally expressed as $\text{CGP}=(\text{Con}\mid \Upsilon)$. Here we look into the resulting route sheets $\Upsilon=\{\mathcal{J},\mathcal{M}\}$ from the previous module. These route sheets outline the production procedure $J=\langle O_1,O_2,\dots,O_{|J|}\rangle$ for each target product, detailing the execution sequence of involved operations. The complete set of operations from all procedures, denoted as $\mathcal{O}_{\mathcal{J}}=\{O_1,O_2,\dots,O_{|\mathcal{O}_{\mathcal{J}}|}\}$, forms a partially ordered set that indicates the inter-dependency relationships among operations. If the precondition of $O_i$, namely the availability of input materials required for $O_i$'s execution, includes the postcondition of $O_j$, which is the output product following the execution of $O_j$, then $O_i$ is \emph{dependent} on $O_j$, denoted as $\text{dep}(O_i,O_j)$. Furthermore, an operation $O_i$ must be executed on a specific machine $M_j$, represented by the relationship $\text{exec}(O_i,M_j)$. 

Using the aforementioned notations, we can represent the underlying \ac{jsp} for the production scenario with $\mathcal{J}$, $\mathcal{M}$, and $\mathcal{O}$. To solve the \ac{jsp}, it is necessary to specify (i) the set of resource constraints $\mathcal{R}=\{(O_i,M_j)\mid O_i\in\mathcal{O}, M_j\in\mathcal{M}, \text{exec}(O_i,M_j)=\text{True}\}$; and (ii) the set of precedence constraints $\mathcal{P}=\{(O_i,O_j)\mid O_i,O_j\in\mathcal{O},\text{dep}(O_i,O_j)=\text{True}\}$. The objectives of the \ac{jsp} may include maximizing throughput, minimizing response time, or balancing resource utilization. As the objective is independent of constraint specification and thus falls outside the scope of this work, we exclude it from the problem definition for succinctness. In summary, we define the constraint generation problem as the task of specifying the resource and precedence constraints $\text{Con}=\{\mathcal{R},\mathcal{P}\}$, given the description of procedures and the actual condition of factories. 

We specify $\text{Con}$ from $\Upsilon$ through \ac{dsl} program verification, adapted from the methodology of Shi~\etal~\cite{shi2024expert}. The \ac{dsl} programs are verified by associating operations with product flows in a reciprocal manner. Product flow indicates the transfer of product flow units among operations, reflecting how one operation influences subsequent ones. The program verifier traverses the \ac{dsl} program in execution order, utilizing the product locality revealed from the actual distribution of operations and product flow units. This process determines the \emph{reachability} and \emph{life cycle} of product flow units, in accordance with the theory of compilation introduced by Aho and Ullman~\cite{aho1972theory}. For the implementation of the verifier, a \ac{pda} with a random access memory is employed to record reachable product flow units as an operation context, defining (\ie, the product is produced by certain operations or the raw material is purchased) and killing (\ie, the product is consumed by certain operations) product flow units at each operation point along the computation. During every transition between operations, the killed products are removed from the memory, and the defined products are added to it. After a product flow unit is killed, the pair of operations that defined it and killed it is added to the set of precedence constraints $\mathcal{P}$. At each operation point, the pair of the operation and the machine specified in its execution configuration is added to the set of resource constraints $\mathcal{R}$. The accepting state of the \ac{pda} is reached if the memory is empty at the end of execution, meaning all products defined in operations are killed by other operations. Alongside the deterministic \ac{pda}-based verifier, We employ state-of-the-art \acp{llm} to track and monitor the two actions of the \ac{pda}, kill and define, through instruction-following in-context learning~\cite{brown2020language,wei2022finetuned}. 

The specified constraints, along with the route sheets, are converted into the input format for the \ac{jsp} solver, specifically using the widely adopted OR-Tools \ac{jsp} solver\footnote{The \ac{jsp} solver is implemented using the OR-Tools library, with documentation available at \url{https://developers.google.com/optimization/scheduling}.}. Within the \ac{jsp} solver framework, the manufacturing scheduling problem is modeled as a job shop environment comprising $M$ machines and $N$ jobs. Each job consists of a series of operations that may have inter-dependencies and require specific machine types, as indicated by precedence and resource constraints, respectively. Each operation must be assigned to a machine and processed within a specified execution duration, so as to fulfill the objective of optimizing the overall manufacturing process~\cite{graham1966bounds,graham1979optimization}. With these parameters, the \ac{jsp} can be effectively solved using constraint programming techniques~\cite{baptiste2001constraint,baptiste2006constraint}. The resulting schedules specify the exact time-machine-operation arrangements for production.

\subsection{Schedule Grounding into Production Plans}\label{subsec:arch-j2p}

The third module converts the schedules generated by the \ac{jsp} solver into interpretable production plans within the factory environment. This module is essential for integrating off-the-shelf \ac{jsp} solvers into the entire architecture, as it involves translating the physical meanings of route sheets and constraints from their \ac{dsl} representations into a format compatible with the solver's input. Consequently, it is necessary to recontextualize the schedules to their physical meanings for practical application. The execution configurations in the grounded production plans must match those in the route sheets, and the execution timing must adhere precisely to the generated schedules. To ensure reliability, this grounding is achieved through symbolic \ac{dsl} program referencing.

The working mechanism of this module can be formally expressed as $\text{SGP}=(\text{Exe}(\mathcal{J},\mathcal{M})\mid \text{JSP}(\text{Con}, \Upsilon))$, where $\text{JSP}(\text{Con}, \Upsilon)$ denotes the output of the \ac{jsp} solver. Meanwhile, $\text{Exe}(\mathcal{J},\mathcal{M})=\langle(O_i,M_j,t_1),\dots, (O_k,M_l,t_{|\text{Exe}(\mathcal{J},\mathcal{M})|})\rangle$ denotes the grounded production plans that meticulously arrange operations and machines within the temporal dimension. The incorporation of timing necessitates that the \ac{dsl} program representation comprises a concurrent mechanism~\cite{chandy1989parallel}. Specifically, an operation must wait to start until all operations producing the intermediate products required by its precondition have been completed. Given that our \ac{dsl}, which adopts the product-flow-centric view, possesses the syntactic feature to model product-based transitive relationships among operations, it seamlessly facilitates the interpretation of the predecessor and successor of a product flow unit as \emph{process locks}, serving as a guardrail for the correctness of the production plans. Furthermore, the \ac{dsl} program verifier introduced in \cref{subsec:arch-c2j} functions as an additional assurance of correctness. It traverses the dependency graph to verify the absence of breakpoints in the product flow along the temporal dimension. With this dual-guardrail mechanism, we can directly reference the \ac{dsl} programs using their semanticless identifiers within the \ac{jsp} solver, thereby benefiting from the determinism of symbolic representations. Subsequently, we can verify the correctness of the resulting production plans.

\begin{figure*}[ht]
    \centering
    \includegraphics[width=\linewidth]{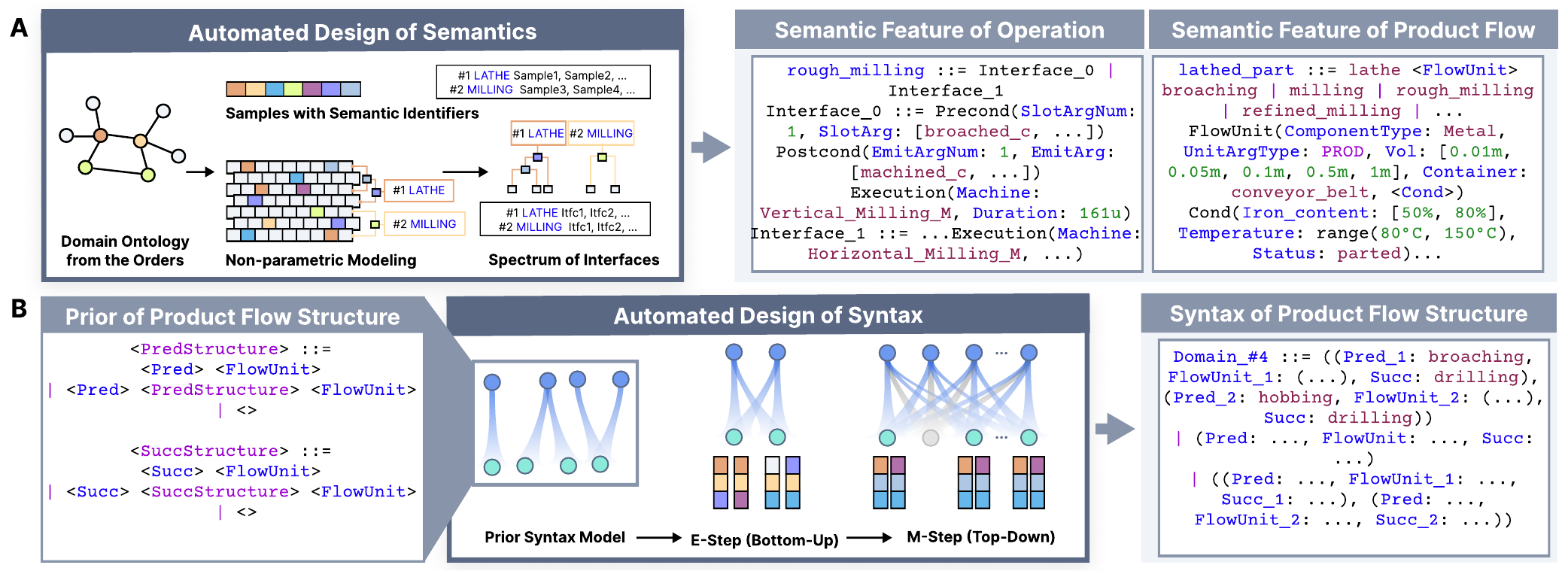}
    \caption{\textbf{Illustration of the algorithms for the automated production scenario adaptation of the architecture.} \textbf{(A)} This diagram illustrates the framework of non-parametric modeling for the automated design of semantic features within both operation-centric and product-flow-centric program view \acp{dsl}. \textbf{(B)} This diagram depicts the framework of the EM algorithm for the automated design of syntactic features within product-flow-centric program view \ac{dsl}.}
    \label{fig:adaptation}
\end{figure*}

\section{Automated Adaptation Algorithm}\label{sec:auto}

In this section, we present the automated production scenario adaptation of the proposed constraint-centric architecture. Initially, we explore the significance of such adaptation within the manufacturing context (\cref{subsec:auto-why}). We then define the problem of adapting the desired architecture by means of \ac{dsl} design (\cref{subsec:auto-problem}; refer to \cref{fig:adaptation}A). Afterwards, we introduce approaches for the automated design of the \ac{dsl} from the operation-centric program view (\cref{subsec:auto-operation}; also refer to \cref{fig:adaptation}A) and the product-flow-centric program view (\cref{subsec:auto-product-flow}; refer to \cref{fig:adaptation}B) respectively.

\subsection{Why Automated Production Scenario Adaptation?}\label{subsec:auto-why}

While it is theoretically possible to automate the entire workflow described in \cref{sec:arch} for generating grounded production plans, a crucial challenge remains: full automation is contingent upon the availability of predefined constraint-centric architectures, \ie, the corresponding \acp{dsl} for constraint specification. However, the origin of these \acp{dsl} poses a problem, as they are not readily available like off-the-shelf \acp{gpl}. In current practices, most \acp{dsl} are manually designed through the collaborative efforts of computer scientists and manufacturing experts, a process that is both time-consuming and labor-intensive. This may be acceptable for specific applications requiring only a single \ac{dsl} library~\cite{voelter2011product,meixner2021domain,hofmann2023knitscript}, as \ac{dsl} design is a \emph{once-and-for-all} endeavor there. Unfortunately, from a broader perspective of the holistic manufacturing community, \acp{dsl} for constraint specification in manufacturing encompass multiple categories of target products, diverse requirements of the \acp{oem}, varied production environments across factories, and an ever-expanding range of overall application scenarios. This scenario specificity implies that the distributions of operations, machines, materials, intermediate products, execution configurations, and inter-dependency relationships vary significantly among different scenarios. Although it is conceivable that we could derive a comprehensive set of language features covering all potential manufacturing scenarios, namely the so-called \emph{one-size-fits-all general architecture}, such an endeavor would result in a system of prohibitively complexity, rendering it intractable for both machine and human end-users.

The highly varied and frequently evolving demands for \acp{dsl} are difficult to meet through human effort alone. Even if we manage to manually craft these \acp{dsl}, the advancement in automated production planning and scheduling would be compromised. This would merely shift human labor from one part of the workflow to another, and even potentially increase the overall labor required. Maintaining a joint cohort of experienced manufacturing experts and computer scientists for \ac{dsl} design at a higher level than the current factory could prove more costly than simply sustaining experts who work at the existing factory level. Consequently, we find ourselves in a dilemma: \acp{gpl}, which are easily accessible, are unsuitable for constraint specification due to their overwhelming complexity, whereas \acp{dsl}, which simplify specialized language features, inherently lack generalizability across different manufacturing scenarios. To address this dilemma, rather than waiting for a universally applicable \ac{gpl} to emerge, a more practical solution might involve automating the design of \acp{dsl} for constraint specification. Therefore, the automated production scenario adaptation of our constraint-centered architecture is an essential requirement to fully unleash its potential of usability for the broader manufacturing community.

\subsection{The Production Scenario Adaptation Problem}\label{subsec:auto-problem}

We conceptualize the problem of the automated adaptation for the constraint-centric architecture within a specified manufacturing scenario as $\text{DAP}=(\{\mathcal{L}_o^*,\mathcal{L}_p^*\}\mid\Gamma')$. The objective is to design a \ac{dsl} with language features accommodating both the operation-centric program view $\mathcal{L}_o^*$ and the product-flow-centric program view $\mathcal{L}_p^*$. The input $\Gamma'$ constitutes a generalized set of original procedures, which may either be novel procedures from a recently established manufacturing scenario or historical procedure data from a long-standing manufacturing scenario. The prior knowledge of operations and product flows, represented by $p(\varphi)$ and $p(\omega)$, includes the fundamental syntax of the field-value structures and the elementary taxonomies, derived according to the general commonsense of manufacturing. Specifically, the problem essentially seeks to fit the joint distribution models $p(\varphi,\phi, \varphi^{\text{prec}},\varphi^{\text{post}}, \varphi^{\text{exec}})$ and $p(\omega,\omega^{\text{pred}},\omega^{\text{succ}},\omega^{\text{prop}},\psi)$, using $\Gamma'$ as the domain-specific corpus, and leveraging the prior knowledge $p(\varphi)$ and $p(\omega)$.

\subsection{Automated Operation-Centric DSL Design}\label{subsec:auto-operation}

The key challenge in the automated design of the operation-centric program view is to aggregate all possible execution contexts for an operation, and then generalize the contexts to the interface. If we keep each use case as one single instance of the interface, which can be in hundreds regarding one operation, the generalization is meaningless. Since there is no prior knowledge about the interface in advance, we develop the algorithm following the idea of non-parametric modeling, \ie, \ac{dpmm}, resulting in flexible identification of interface instances.  

As we must handle information coming in different granularities, from interface structures (\ie, the main body describing the execution of a specific operation within a production procedure) to values of parameters (\ie, the configuration of machines and consumption of materials within an operation), we choose to model the operations in a hierarchical fashion. Compared with the flattened spectral clustering approach developed by Shi~\etal~\cite{shi2024autodsl}, which compresses all information of an operation into an embedding vector, our modeling is competent in considering information at different levels comprehensively. We carefully adopt the prerequisite that the interface is generated subject to the operation, preconditions, postconditions, and execution configurations are generated subject to the interface, and the value of configuration parameters, denoted as $\varphi^{\text{exec-v}}$, are generated subject to their corresponding fields. Thus, we have the model
\begin{equation}
    \begin{aligned}
        &p(\varphi,\phi, \varphi^{\text{prec}},\varphi^{\text{post}}, \varphi^{\text{exec}}, \varphi^{\text{exec-v}})\\
        =&p(\varphi^{\text{exec-v}}\mid\varphi,\phi,\varphi^{\text{exec}}) p(\varphi^{\text{exec}}\mid\varphi,\phi)p(\varphi^{\text{prec}}\mid\varphi,\phi)\\
        \quad\;&p(\varphi^{\text{post}}\mid\varphi,\phi)p(\phi\mid\varphi)p(\varphi).
    \end{aligned}
\end{equation}
Within each iteration of the \ac{dpmm} process, we sample the variables level-by-level. Since the structures of preconditions, postconditions, and the selection of devices and configuration parameters are discrete, we sample them directly from the \ac{dp}. As permissible values of parameters can be discrete, \eg, an array of specific values, common in temperature preparation; continuous, \eg, an interval with minimum and maximum values, common in power setting; or mixed, \eg, an array of specific values with random perturbations around the mean, common in tooling accuracy control, we conduct the sampling by integrating \ac{gp} with \ac{dp}, obtaining
\begin{equation}
    \begin{aligned}
        &\varphi^{\text{exec-v}}\mid\varphi,\phi, \varphi^{\text{exec}}\\&\sim DP(\alpha,H(\varphi^{\text{exec}}), \phi, \varphi)\times GP(m,K),
    \end{aligned}
\end{equation}
where $\alpha$, $H$, $m$, and $K$ are corresponding hyperparameters. 

While clustering similar interface instances aggregates target operations, there may remain redundant interfaces due to minor discrepancies. These discrepancies often arise from differences in parameter values or naming conventions that do not fundamentally alter the operation's functionality. To alleviate such redundancies, we implement a unification process for the interfaces. Specifically, interface instances associated with the same operation are considered equivalent if they have the same number of slots and emits, and share the same fields in their execution configuration parameters. By abstracting away differences in parameter values and names, we unify these interfaces into a single, generalized interface, akin to the algorithm proposed by Martelli~\etal~\cite{martelli1982efficient}. Unification enhances the generality of the operation-centric program view by consolidating functionally-identical interfaces, maintaining a concise and representative set of operations.

\subsection{Automated Product-Flow-Centric DSL Design}\label{subsec:auto-product-flow}

One of the primary challenges in the automated design of the product-flow-centric program view lies in selecting proper descriptive properties of a product flow unit component. There exists false positive cases, where properties are attributed to components with the same semantic identifier but in different phases, \eg, we consider Aluminum Alloy with the property dimensions when it comes in flake and with the property volume when it comes in powder. There also exists false negative cases, where exact same components are regarded as different ones due to different reference names, \eg, Aluminum Sheet, 6061 Alloy Plate, and Metal Sheet can refer to the same thing. To alleviate false positive and false negative results, we discard the design choice of the interface in the operation-centric view, which tends to cover the possibly richest context, and thereby have the non-parametric model
\begin{equation}
    \begin{aligned}
        &p(\omega,\omega^{\text{pred}},\omega^{\text{succ}},\omega^{\text{prop}},\omega^{\text{prop-v}})\\=&p(\omega^{\text{prop-v}}\mid\omega^{\text{prop}},\omega) p(\omega^{\text{prop}}\mid\omega)
        \\\quad\;&p(\omega^{\text{pred}}\mid\omega)p(\omega^{\text{succ}}\mid\omega)p(\omega),
    \end{aligned}
\end{equation}
where $\omega^{\text{prop-v}}$ denotes the values of property parameters. The challenges in building up this model and the corresponding strategies are similar to those in \cref{subsec:auto-operation}.

The other primary challenge in the automated design of the product-flow-centric program view is constructing the model of $\psi$. This model is not captured by $p(\omega,\omega^{\text{pred}},\omega^{\text{succ}},\omega^{\text{prop}},\omega^{\text{prop-v}})$, yet it remains crucial for completing the design of $\mathcal{S}_p$. Leveraging existing knowledge on programming language design, our method utilizes a bidirectional optimization strategy to formulate the most appropriate flow-structure-syntax $\psi^*\in\mathcal{S}_p^*$ of the target \acp{dsl}, ensuring that it compactly satisfies the characteristics dictated by $\Gamma'$. Inspired by the methodology introduced in Shi~\etal~\cite{shi2024autodsl}, the algorithm utilizes an \ac{em} framework, where the E-Step abstracts syntax from $\Gamma'$ and the M-Step derives syntax from programming language principles.

The algorithm models latent syntactic constraint assignments $\mathcal{Z}=\{z_1,\dots,z_{|\Gamma'|}\}$ for each procedure $\gamma\in\Gamma'$. A filter set $\Theta=\{\theta_1,\dots,\theta_{|\mathcal{Z}|}\}$, is designed to determine if a segment of procedure description, \ie, a local set of product flow units with predecessor and successor operations, aligns with the $N$-predecessors-to-$M$-successors relationships within the product flow, coming with the belief function $p(\Theta|\psi)$. The observational likelihood is computed as
\begin{equation}
    p(\Gamma'\mid\mathcal{Z},\Theta)=\prod^{\Gamma'}_{i=1}p(\gamma_i\mid z_i,\theta_{z_i}).
\end{equation}
Hence, the overall joint distribution of the model is given by
\begin{equation}
    p(\Gamma',\mathcal{Z},\Theta\mid\psi)=p(\Gamma'\mid\mathcal{Z},\Theta)p(\mathcal{Z}\mid\psi)p(\Theta\mid\psi).
\end{equation}

Programming language designers leverage a general set of syntactic production rules as the prior $p(\mathcal{Z}\mid\psi)$ for syntax specification. Following this common practice, we initialize $\psi$ with the recursive grammar
\begin{equation}
    \begin{aligned}
        &\texttt{PredS} ::= \langle\omega^{\text{pred}}\rangle\;\langle\omega^{\text{prop}}\rangle\mid\langle\omega^{\text{pred}}\rangle\;\texttt{PredS}\;\langle\omega^{\text{prop}}\rangle\mid\langle\;\rangle,\\
        &\texttt{SuccS} ::= \langle\omega^{\text{succ}}\rangle\;\langle\omega^{\text{prop}}\rangle\mid\langle\omega^{\text{succ}}\rangle\;\texttt{SuccS}\;\langle\omega^{\text{prop}}\rangle\mid\langle\;\rangle,
    \end{aligned}
\end{equation}
where $\langle\omega^{\text{pred}}\rangle\;\texttt{PredS}\;\langle\omega^{\text{prop}}\rangle$ and $\langle\omega^{\text{succ}}\rangle\;\texttt{SuccS}\;\langle\omega^{\text{prop}}\rangle$ are recursion bodies. This recursive grammar accommodates hypotheses involving arbitrary $N$-predecessors-to-$M$-successors relationships, naturally beginning with the simplest linear structure and progressively increasing in complexity. Additionally, we construct the prior belief function $p(\Theta\mid\psi)$ with a series of sliding-window-based filters $f: \Gamma' \mapsto \mathbb{R}$. This approach provides a relaxed lower bound for predicting the existence of an atomic product-flow structure. 

In each E-Step, we obtain the posterior of latent variables $p(\mathcal{Z}\mid\Gamma',\Theta,\psi)$ applying Bayes' theorem, which is implemented by scanning the filters over all procedure descriptions in $\Gamma'$. To note, as the spaces of prior and observation are not intractably large, we simply employ the naive version of E-Step without variational approximations.

In each M-Step, we first maximize the coverage of the sampled atomic structure $\psi$ by maximizing
\begin{equation}
    \mathcal{Q}(\hat{\Theta},\Theta)=\mathbb{E}_{\mathcal{Z}\mid\Gamma',\Theta}\big[\log p(\Gamma',\mathcal{Z},\hat{\Theta}\mid\psi)\big],
\end{equation}
where $\hat{\Theta}$ is the updated $\Theta$, resulting in the structural change of $\psi$. These two steps alternate iteratively until convergence, ensuring the syntactic features are aligned with the scenario.

\section{Experimental Setups}\label{sec:exp}

In this section, we describe the experimental setups of this study. We first introduce the datasets for experimentation (\cref{subsec:exp-data}) and the baseline approaches for evaluation  (\cref{subsec:exp-baseline}). Afterwards, we describe the protocols for the five experiments, including the complete pipeline experiment (\cref{subsec:exp-pipeline}), three experiments validating the three modules of our constraint-centric architecture (\cref{subsec:exp-o2c,subsec:exp-c2j,subsec:exp-j2p}), and the production scenario adaptation experiment (\cref{subsec:exp-adapt}).

\subsection{Datasets for Experimentation}\label{subsec:exp-data}

The scarcity of datasets that closely replicate real-world manufacturing environments significantly hinders the evaluation of our constraint-centric architecture. To bridge this gap, we propose the augmentation of existing datasets with synthetic data, ensuring the retention of realistic elements in the process. This approach involves the utilization of ten classical \acp{jsp} sourced from well-established \ac{or} literature~\cite{taillard1993benchmarks,adams1988shifting,applegate1991computational,bierwirth1995generalized,balas1998guided,carlier1989algorithm,jain1999deterministic,nowicki1996fast,sabuncuoglu1999job,storer1992new}. These \acp{jsp}, originally comprising only machine IDs and durations, serve as the foundation for our dataset enhancement.

To transform abstract \ac{jsp} descriptions into comprehensive datasets, we employ a cutting-edge \ac{llm}\footnote{We use the OpenAI \texttt{GPT-4o} model for this purpose.}, which extends the sparse \ac{jsp} data by generating detailed production procedure descriptions and semi-structured route sheets in \ac{nl}. This transformation aligns the synthetic data with the style and complexity of realistic production procedures, thereby enhancing the practical value of the augmented dataset.

The augmentation process begins with a dependency graph traversal to ascertain the global dependency set across an array of device types. This step is crucial for understanding the inter-dependency relationships and operation sequences within the manufacturing setup. Subsequently, a mapping arrangement is established between the machine IDs and the corresponding devices. Each \ac{jsp} is treated as a distinct \emph{production scenario} requiring a \ac{dsl} for accurate constraint specification. The dependency set for each machine arrangement is meticulously configured to be a superset of the \ac{jsp}'s dependency set, adhering to an initial assumption of ordered and linear job dependencies. Non-monotonic dependencies (\ie, circular steps) are identified and eliminated to prevent operational conflicts.

Following the establishment of device mappings and dependency configurations, the generation of synthetic data begins. This phase involves the specification of materials, products, and the execution configurations of devices. Through this data synthesis process, we obtain a dataset that not only reflects the complexity of real-world manufacturing tasks but also serves as a controllable probe for the three respective modules within our architecture and its baseline counterparts. This supports our experimental setups, enabling a thorough evaluation of our system's performance across various scenarios. It illustrates that our approach effectively harnesses the advantages of both \ac{genai} techniques and \ac{dsl}-based structural representation, thereby achieving a balance between the capability of generation and the guardrail of reliability~\cite{shi2024constraint}.

\subsection{Baseline Approaches}\label{subsec:exp-baseline}

To establish a robust evaluation framework, we implement two alternative approaches based on state-of-the-art methods for formalizing \ac{nl}-described optimization problems with \ac{llm} prompt engineering techniques~\cite{xiao2023chain,ahmaditeshnizi2024optimus}. These approaches, termed \textsc{Multi-Stage-LLM (MSL)} and \textsc{Two-Stage-LLM (TSL)}, serve as baselines to benchmark the utility of our proposed constraint-centric architecture (Ours).

The MSL approach adopts a three-module-sequence that mirrors the structural alignment of our constraint-centric architecture, facilitating a direct comparative analysis. Initially, the workflow transforms \ac{nl}-based descriptions or semi-structured route sheets into fully structured route sheets. Subsequently, the fully structured route sheets are converted into matrices, in a format matchable for the \ac{jsp} solver. The final module integrates the schedules generated by the \ac{jsp} solver with the fully structured route sheet, and then grounds these schedules into interpretable production plans. All of these three modules are implemented by \ac{llm} prompt engineering. \footnote{We implement these \acp{llm} with the OpenAI \texttt{GPT-4o} model.}

In contrast to the MSL, the TSL approach simplifies the workflow by reducing the number of transformation stages, potentially increasing computational efficiency but at the risk of reduced fidelity in the constraint translation process. The first stage bypasses the separate structuring step of the route sheet and directly converts the \ac{nl}-based descriptions or semi-structured route sheets into \ac{jsp} solver formatted matrices. The second stage involves the integration of the \ac{jsp}-generated schedules with the input procedures, and then grounds these integrated schedules into finalized production plans. Both modules are implemented by \ac{llm} prompt engineering.

\subsection{Protocol for the Complete Pipeline Evaluation}\label{subsec:exp-pipeline}

The major objective of this experiment is to validate the utility of our constraint-centric architecture within realistic manufacturing scenarios. This experiment is critical in demonstrating the practical utility of our approach as the primary outcome of this study. The experimental protocol involves a comparative analysis of Ours against two baseline approaches, MSL and TSL. These methods are evaluated across ten realistic manufacturing scenarios as detailed in \cref{subsec:exp-data}. To ensure a fair comparison, the base \ac{llm} model for all three approaches is kept identical. The input to these pipelines consists of procedures described either in \ac{nl} or as semi-structured route sheets. We keep the proportion of both input forms identical throughout the experiment. The output from each system is formatted into JSON-style scripts, which align with the script formats used for \ac{cnc} system execution, allowing for a standardized method of comparison.

The evaluation of these scripts is conducted using the \ac{emkvp} and \ac{bleu} metrics, which are well-established in the literature for assessing the quality of procedural knowledge representation and information retrieval~\cite{schutze2008introduction,papineni2002bleu}, against the ground truth production plans derived from the synthetic dataset under the supervision of manufacturing experts. Given that direct comparisons using cosine similarity score across entire sentences could lead to inaccuracies due to semantic discrepancies in similar-looking instructions, we adopt a more granular approach. By converting all results into a standardized JSON-style format, comparisons are made between field-value pairs rather than entire sentences using the \ac{emkvp} metric. Meanwhile, procedure-level consistencies are assessed through the \ac{bleu} metric. This hybrid method effectively alleviates concerns related to the metrics by focusing either only on the precision and accuracy of specific data elements or only on overall textual similarity. Furthermore, the results are quantitatively analyzed using three specific variants of the \ac{emkvp} metric: Precision, Recall, and F1 Score. \ac{emkvp}-Precision measures the proportion of correct field-value pairs among all specified pairs, providing insight into the accuracy of the resulting production plans. \ac{emkvp}-Recall assesses the proportion of correctly specified field-value pairs out of all pairs that should have been specified, reflecting the completeness of the information captured. Lastly, \ac{emkvp}-F1 Score combines both precision and recall to offer a balanced view of overall performance.

\subsection{Protocol for the Constraint Abstraction Evaluation}\label{subsec:exp-o2c}

The primary objective of this experiment is to validate the effectiveness of the constraint abstraction module (CAM) within our constraint-centric architecture. This module is crucial for ensuring that ambiguities in \ac{nl} parsing are managed effectively and that the precision required for fine-grained procedural knowledge representation is maintained. By isolating this component, we aim to demonstrate the indispensable role of \acp{dsl} as mechanisms that guide and regulate the behavior of \acp{llm} within our system. The experimental setup involves using identical input data as employed in the complete pipeline experiment. The outputs generated from this input are fully-structured route sheets, which are subsequently transformed into JSON format for consistency and ease of analysis. This standardized format allows for direct comparison between the outputs from the CAM of Ours (Ours-CAM) and those from the first module of MSL (MSL-I). Similar to \cref{subsec:exp-pipeline}, the evaluation metrics include \ac{bleu}, \ac{emkvp}-Precision, \ac{emkvp}-Recall, and \ac{emkvp}-F1. These metrics are chosen for their ability to quantitatively measure the consistency between the resulting route sheets and the groundtruth fully-structured route sheets derived from synthetic data under the supervision of manufacturing experts.

\subsection{Protocol for the Constraint Generation Evaluation}\label{subsec:exp-c2j}

The primary objective of this experiment is to validate the utility of the constraint generation module (CGM) within our constraint-centric architecture. This module is significant for ensuring the correctness of the constraints for \ac{jsp} formulations and for converting these constraints into a format compatible with \ac{jsp} solvers without any loss of information.

We have designed two versions of this experiment. The first version assesses the CGM in Ours (Ours-CGM) and the second module in MSL (MSL-II) to highlight the critical role of \ac{dsl} program verification over the dual-program-view representation. This approach is instrumental in capturing the relationships between operations and machines, as well as among operations, in both Ours and MSL. The second version tests the integration of the first two modules, CAM \& CGM, to explore the trade-offs involved in using two sequential modules with accumulated error transmission between them (\ie, Ours-CAM-CGM and MSL-I-II) versus employing a single module without an explicit fully-structured route sheet as an intermediate result to work with (\ie, TSL-I). For the CGM-only version, we use the groundtruth fully-structured route sheet as the input for both Ours-CGM and MSL-II to eliminate accumulated errors and truly isolate the CGM. For the CAM \& CGM version, we use the same input as in the complete pipeline experiment for Ours-CAM-CGM, MSL-I-II, and TSL-I. The output for both versions is the set of constraint generation matching the input format of the \ac{jsp} solver.

Given that the scheduling optimization method used in the \ac{jsp} solver is deterministic, it is unnecessary to assess the correctness of the schedules generated by the \ac{jsp} solver. Therefore, we adhere to established literature on benchmarking the formalization of \ac{nl}-described optimization~\cite{ramamonjison2023nl4opt}, incorporating three evaluation metrics: constraint-level accuracy (Constraint-Acc), compiler error rate (Compiler-ER), and runtime error rate (Runtime-ER). Constraint-Acc is calculated using the \ac{iou} metric between the specified constraints and the groundtruth constraints, encompassing both resource and precedence constraints. This metric indicates the consistency between the intended procedures and the interpretations by the pipelines. Both Compiler-ER and Runtime-ER concern the interactions between the pipeline and the \ac{jsp} solver. The former captures the proportion of testing procedures that fail to compile in the \ac{jsp} solver, potentially due to syntactic-level issues in the formatted set of constraints provided as input. The latter measures the proportion of testing procedures encountering errors during the execution of the \ac{jsp} solver, which are caused by semantic-level errors in the formatted set of constraints, such as internal logic errors, unsolvable models, or non-linear constraints.

\subsection{Protocol for the Schedule Grounding Evaluation}\label{subsec:exp-j2p}

The major objective of this experiment is to validate the usefulness of the schedule grounding module (SGM) within our constraint-centric architecture. This module is essential for ensuring the accurate recovery of fine-grained procedure knowledge from the semantically void schedules produced by the \ac{jsp} solver. By isolating this component, we aim to demonstrate the pivotal role of the \ac{dsl} dual program view in managing the concurrent programming nature of the grounded production plans. We input the groundtruth specified set of constraints into the \ac{jsp} solver to obtain the schedule. This schedule, which remains consistent across different runs, serves as the input for the SGM in Ours (Ours-SGM), the third module of MSL (MSL-III), and the second module of TSL (TSL-II). The outputs generated from this input are grounded production plans, which are subsequently converted into JSON format for consistency and ease of analysis. This standardized format facilitates direct comparison across the three pipelines. Consistent with \cref{subsec:exp-pipeline}, the evaluation metrics include \ac{bleu}, \ac{emkvp}-Precision, \ac{emkvp}-Recall, and \ac{emkvp}-F1. These metrics are selected for quantitatively assessing the consistency between the resulting production plans and the same ground truth used in the complete pipeline experiment. 

\subsection{Protocol for the Automated Adaptation Evaluation}\label{subsec:exp-adapt}

The primary objective of this meta-study experiment, building upon the previous experiments, is to evaluate the scalability of our constraint-centric architecture across various manufacturing scenarios. This characteristic is central to the broader impact of our architecture on the entire manufacturing community, as it results from a trade-off between compromised generality and enhanced domain-specificity---the generalizability is thus alternatively amortized by the automated production scenario adaptation capability, as discussed in \cref{subsec:auto-why}.

This meta-study comprises two components. The first involves observing the convergence of our algorithms for automated adaptation across the ten selected production scenarios. Each of the ten testing groups is initialized without any external prior knowledge, except for what is already integrated into the algorithms, such as the syntactic prior of recursion. The algorithms in these ten groups are expected to perform uniformly well, without case-specific bias. With these automatically designed \acp{dsl} tailored for each scenario, the second component involves validating the utility of the scenario-adapted architecture driven by the corresponding \ac{dsl} for each scenario. This includes all previous experiments, such as the complete pipeline experiment and the three experiments on the individual modules of our architecture. It is expected that these experiments will not exhibit significant differences across the various scenarios. In addition, we utilize the \ac{vmr} metric, which is suitable for one-approach-multiple-domain evaluation~\cite{shi2023perslearn}, to evaluate the quantitative results of Ours, MSL, and TSL across the ten scenarios. This metric, where higher values are more desirable, reflects the simultaneous achievement of both outstanding and consistent performance across the different scenarios.

\begin{figure*}[t!]
    \centering
    \includegraphics[width=\linewidth]{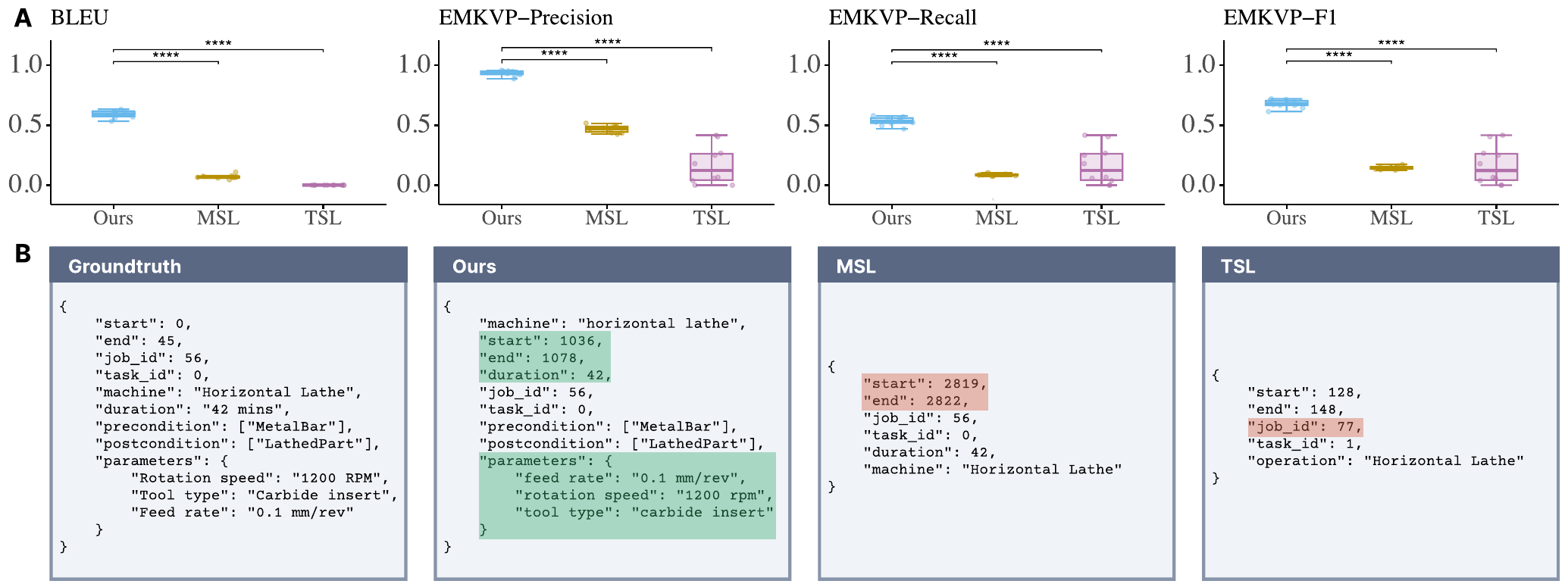}
    \caption{\textbf{Results of the complete pipeline evaluation.} \textbf{(A)} Comparison of Ours with MSL and TSL across four evaluation metrics over ten scenarios. \textbf{(B)} Showcases of the grounded production plans generated by Ours, MSL, and TSL, respectively.}
    \label{fig:res-pipeline}
\end{figure*}

\begin{figure*}[ht]
    \centering
    \includegraphics[width=\linewidth]{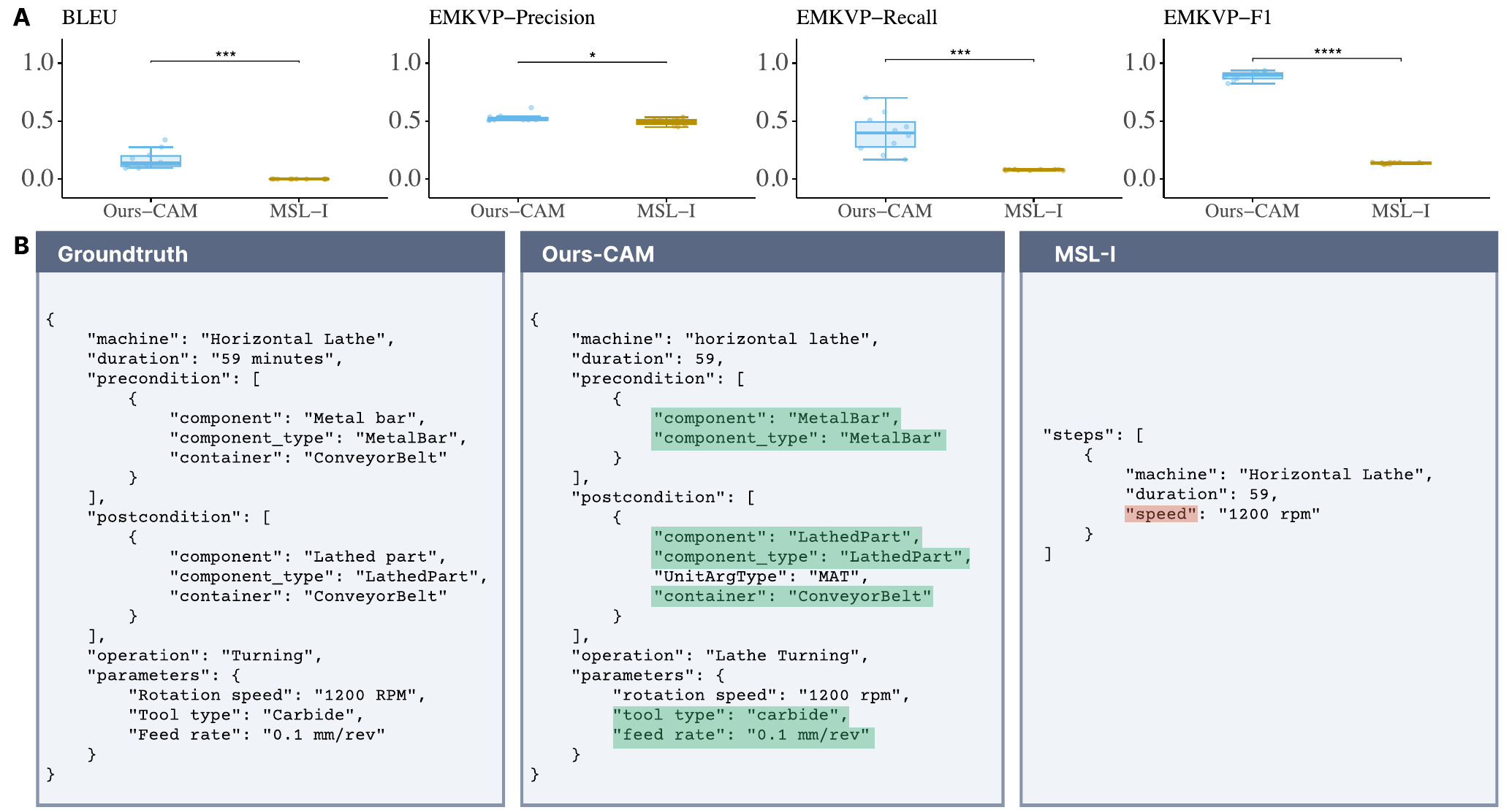}
    \caption{\textbf{Results of the constraint abstraction evaluation.} \textbf{(A)} Comparison of Ours-CAM with MSL-I across four evaluation metrics over ten scenarios. \textbf{(B)} Showcases of the fully-structured route sheets generated by Ours-CAM and MSL-I, and TSL, respectively.}
    \label{fig:res-o2c}
\end{figure*}

\begin{figure*}[ht]
    \centering
    \includegraphics[width=\linewidth]{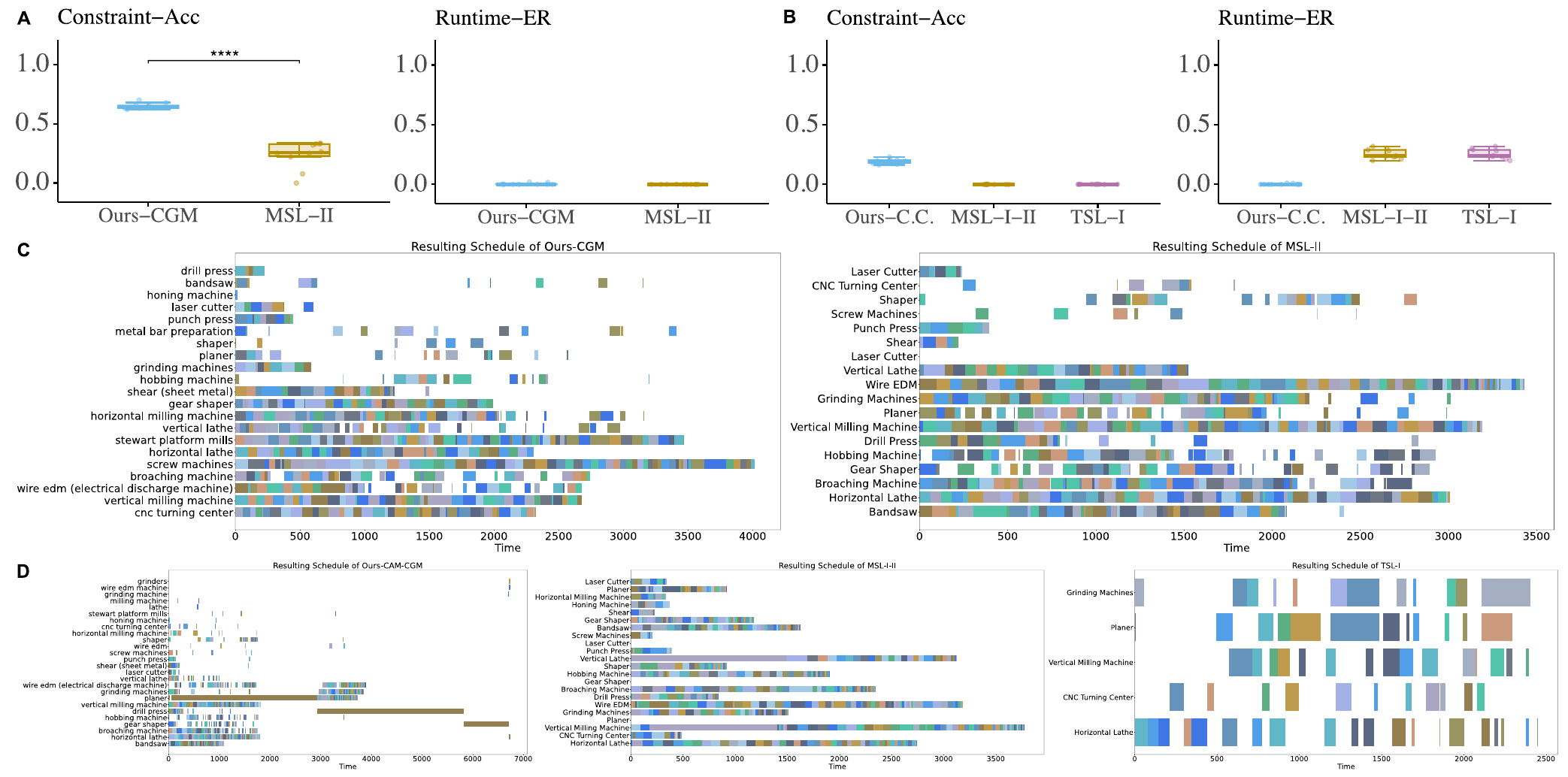}
    \caption{\textbf{Results of the constraint generation evaluation.} \textbf{(A)} Comparison of Ours-CGM with MSL-II across the evaluation metrics Constraint-Acc and Runtime-ER over ten scenarios, within the isolated version of the experiment. Results for Compiler-ER are not visualized because the error cases are minor and consistent across the baseline approaches. Consequently, we have omitted the corresponding plots for the sake of brevity. This design choice aligns with the discussion by Xiao~\etal~\cite{xiao2023chain}. \textbf{(B)} Comparison of Ours-CAM-CGM with MSL-I-II and TSL-I across two evaluation metrics over ten scenarios, within the incorporated version of the experiment. \textbf{(C)} Gantt chart visualizations of the \ac{jsp}-solver-generated schedules, derived from the specified constraints, as generated by Ours-CGM and MSL-II, respectively. The Gantt chart visualization effectively illustrates the differences in modeling fidelity of the \ac{jsp} specification across three approaches. The y-axis represents the various machine types involved in production, revealing that schedules derived from the baseline approaches utilize fewer machines, primarily due to the absence of resource constraints. The x-axis depicts the total production duration, showing that the baseline approaches generate shorter time horizons, likely resulting from omissions in the specification of steps, operations, or machines---a consequence of lacking both resource and precedence constraints. The occupied area within the chart indicates the total machine occupation time. Notably, our approach's schedule contains more unoccupied areas, demonstrating that more machines must remain idle due to procedural requirements---this clearly illustrates that precedence constraints are modeled with greater precision in Ours compared to the baselines. \textbf{(D)} Gantt chart visualizations of the \ac{jsp}-solver-generated schedules, derived from the specified constraints, as generated by Ours-CAM-CGM, MSL-I-II, and TSL-I, respectively. Applying the same criteria as (C), the \ac{jsp} specification modeled by the two baseline approaches demonstrates reduced fidelity, primarily attributable to their failure to incorporate resource constraints and precedence constraints.}
    \label{fig:res-c2j}
\end{figure*}

\section{Results and Discussions}\label{sec:res}

In this section, we analyze the results of the five experiments described in \cref{sec:exp} and discuss the insights revealed by them, including the complete pipeline experiment (\cref{subsec:res-pipeline}), three experiments validating the utilities of the three modules of our constraint-centric architecture (\cref{subsec:res-o2c,subsec:res-c2j,subsec:res-j2p}), and the production scenario adaptation experiment (\cref{subsec:res-adapt}).

\subsection{The Complete Pipeline Evaluation}\label{subsec:res-pipeline}

Through paired samples t-test, we find that Ours significantly outperforms the alternative approaches MSL and TSL across the four evaluation metrics (Ours outperforms MSL, measured by \ac{bleu}: $t(18)=47.448, \mu_d<0, p<.0001$; measured by \ac{emkvp}-Precision: $t(18)=39.143, \mu_d<0, p<.0001$; measured by \ac{emkvp}-Recall: $t(18)=39.528, \mu_d<0, p<.0001$; measured by \ac{emkvp}-F1: $t(18)=46.215, \mu_d<0, p<.0001$; Ours outperforms TSL, measured by \ac{bleu}: $t(18)=60.806, \mu_d<0, p<.0001$; measured by \ac{emkvp}-Precision: $t(18)=15.003, \mu_d<0, p<.0001$; measured by \ac{emkvp}-Recall: $t(18)=7.071, \mu_d<0, p<.0001$; measured by \ac{emkvp}-F1: $t(18)=9.891, \mu_d<0, p<.0001$; see \cref{fig:res-pipeline}A). These comparisons demonstrate the suitability of our architecture for constraint specification. In addition, we find that the baseline approach equipped with an explicit fully-structured route sheet as an intermediate workspace (\ie, MSL) outperforms its counterpart without such workspace (\ie, TSL) (measured by \ac{bleu}: $t(18)=13.612, \mu_d<0, p<.0001$; measured by \ac{emkvp}-Precision: $t(18)=5.873, \mu_d<0, p<.0001$; see \cref{fig:res-pipeline}A). Looking into the resulting production plans, we find that Ours effectively captures fine-grained execution configurations, such as the \emph{``feed rate''} and \emph{``tool type''}. In contrast, MSL partially fails to capture this information, and TSL fails entirely. Among the three approaches, only Ours accurately maintains the consistency between start time, end time, and duration. The pure \ac{llm}-based counterparts even make errors in this aspect, and TSL generates production plans that are \emph{irrelevant}. These observations are illustrated by the examples presented in \cref{fig:res-pipeline}B. 

\subsection{The Constraint Abstraction Evaluation}\label{subsec:res-o2c}

Through paired samples t-test, we find that Ours-CAM significantly outperforms the alternative approach MSL-I across the four evaluation metrics (measured by \ac{bleu}: $t(18)=6.487, \mu_d<0, p<.0001$; measured by \ac{emkvp}-Precision: $t(18)=2.481, \mu_d<0, p<.05$; measured by \ac{emkvp}-Recall: $t(18)=7.342, \mu_d<0, p<.0001$; measured by \ac{emkvp}-F1: $t(18)=28.851, \mu_d<0, p<.0001$; see \cref{fig:res-o2c}A). Notably, the \ac{emkvp}-Precision of MSL-I may be distorted to a certain extent due to its correspondingly low recall---many fields remain unspecified, thus significantly limiting the set of field-value pairs available for evaluation and excluding a substantial number of false negative samples from consideration. These comparisons demonstrate the capability of Ours-CAM for \ac{nl} parsing and fine-grained procedural knowledge representation. Upon examining the resulting route sheets, we observe that Ours-CAM effectively captures fine-grained execution configurations, surpassing MSL-I in this regard. This achievement lays a robust foundation for subsequent processing. These observations are exemplified by the cases presented in \cref{fig:res-o2c}B.

\begin{figure*}[ht]
    \centering
    \includegraphics[width=\linewidth]{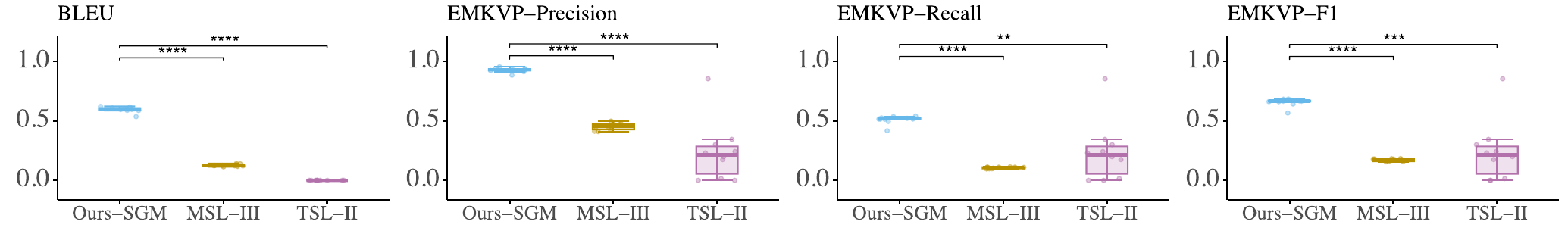}
    \caption{\textbf{Results of the schedule grounding evaluation.} This figure presents the comparison of Ours-SGM with MSL-III and TSL-II across four evaluation metrics over ten scenarios.}
    \label{fig:res-j2p}
\end{figure*}

\begin{figure*}[ht]
    \centering
    \includegraphics[width=\linewidth]{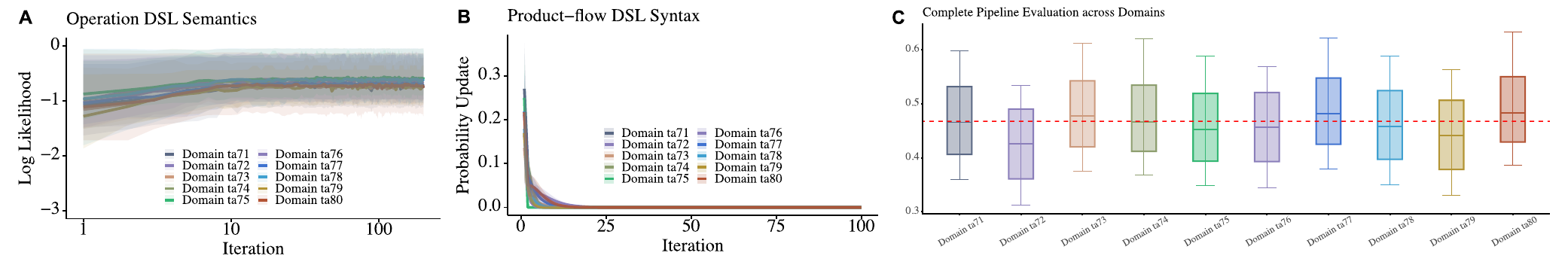}
    \caption{\textbf{Results of the production scenario adaptation evaluation.} \textbf{(A)} Convergence curve of the non-parametric model for the automated design of semantic features within operation-centric program view \acp{dsl}. \textbf{(B)} Convergence curve of the EM algorithm for the automated design of syntactic features in product-flow-centric program view \ac{dsl}. \textbf{(C)} Comparison of Ours with MSL and TSL on ten scenarios respectively integrating the four evaluation metrics.}
    \label{fig:res-adapt}
\end{figure*}

\subsection{The Constraint Generation Evaluation}\label{subsec:res-c2j}

In the isolated version of the experiment, through paired samples t-test, we find that Ours-CGM significantly outperforms MSL-II on the Constraint-Acc metric ($t(18)=11.072, \mu_d<0, p<.0001$; see \cref{fig:res-c2j}A). These results support that Ours-CGM excels constraint generation through \ac{dsl} program verification over the dual program views. 

In the incorporated version of the experiment, through paired samples t-test, we find that Ours-CAM-CGM significantly outperforms the alternative approaches MSL-I-II and TSL-I across the three evaluation metrics (Ours-CAM-CGM outperforms MSL-I-II, measured by Constraint-Acc: $t(18)=30.201, \mu_d<0, p<.0001$; measured by Runtime-ER: $t(18)=-19.562, \mu_d<0, p<.0001$; Ours-CAM-CGM outperforms TSL-I, measured by Constraint-Acc: $t(18)=30.201, \mu_d<0, p<.0001$; measured by Runtime-ER: $t(18)=-19.562, \mu_d<0, p<.0001$; see \cref{fig:res-c2j}B). These comparisons suggest that, despite the accumulated errors transmitted between the sequential modules of Ours-CAM-CGM, the benefits conferred by the explicit, structural intermediate workspace architecture outweigh the drawbacks of the cumulative errors, which are mitigated by TSL-I. An examination of the schedules produced by the \ac{jsp} solvers reveals that the baseline approaches fail to accurately specify neither resource nor precedence constraints. This inadequacy results in schedules characterized by a high rate of task loss and unreliable dependency relationships, as depicted in the Gantt charts displaying decreased thread rows and job blocks (see \cref{fig:res-c2j}C for Ours-CGM \vs MSL-II; see \cref{fig:res-c2j}D for Ours-CAM-CGM \vs MSL-I-II \vs TSL-I). The higher Runtime-ER observed in the baseline models corroborates this issue, as evidenced by the substantial presence of \emph{``undefined''} and \emph{``null''} values in their formatted constraints. These findings substantiate the rationale underpinning the design of our \ac{dsl}-based constraint-centric architecture against pure \acp{llm}.

\subsection{The Schedule Grounding Evaluation}\label{subsec:res-j2p}

Through paired samples t-test, we find that Ours-SGM significantly outperforms the alternative approaches MSL-III and TSL-II across the four evaluation metrics (Ours-SGM outperforms MSL-III, measured by \ac{bleu}: $t(18)=58.332, \mu_d<0, p<.0001$; measured by \ac{emkvp}-Precision: $t(18)=42.726, \mu_d<0, p<.0001$; measured by \ac{emkvp}-Recall: $t(18)=36.001, \mu_d<0, p<.0001$; measured by \ac{emkvp}-F1: $t(18)=43.236, \mu_d<0, p<.0001$; Ours-SGM outperforms TSL-II, measured by \ac{bleu}: $t(18)=79.089, \mu_d<0, p<.0001$; measured by \ac{emkvp}-Precision: $t(18)=8.705, \mu_d<0, p<.0001$; measured by \ac{emkvp}-Recall: $t(18)=3.422, \mu_d<0, p<.01$; measured by \ac{emkvp}-F1: $t(18)=5.273, \mu_d<0, p<.0001$; see \cref{fig:res-j2p}A). These comparisons highlight the effectiveness of Ours-SGM in schedule grounding, particularly through the dual program view of the \ac{dsl} in managing the concurrent programming nature of grounded production plans. In contrast, even when provided with groundtruth schedules, the baseline approaches perform poorly in maintaining consistency among start time, end time, and duration, let alone their effectiveness on addressing the issue of occupation caused by product transitions.

\subsection{The Automated Adaptation Evaluation}\label{subsec:res-adapt} 

In the first component of this meta-study, we present the trends observed from the behaviors of the automated design algorithms. The automated design algorithm of both operation-centric and product-flow-centric program view \ac{dsl} semantics converges on ten scenarios respectively, as illustrated by the likelihood curve yielded by the non-parametric model in \cref{fig:res-adapt}A. The automated design algorithm of product-flow-centric program view \ac{dsl} syntax converges on ten scenarios respectively, as illustrated by the likelihood curve generated by the \ac{em} in \cref{fig:res-adapt}B. Given the case-specificity, the automated design algorithms effectively tailor the resulting \acp{dsl} to align closely with the characteristics of their respective scenarios. These algorithms adeptly capture the unique language features inherent to each scenarios, preserve the common ones, and prune out those deemed unnecessary.

In the second component of this meta-study, the null hypothesis posits that the performance of Ours in uncorrelated with the choice of scenario. Utilizing the Kruskal-Wallis H-Test, we determine that the null hypothesis should be accepted based on the complete pipeline experiment ($H(9)=2.605,p=.978$; see \cref{fig:res-adapt}C). This indicates a lack of evidence supporting a correlation between performance and scenario. Furthermore, we observe that the results from Ours demonstrate a high mean and low variance, resulting in a trend where the \ac{vmr} of Ours significantly surpasses that of TSL but does not significantly exceed that of MSL. This is because a substantial portion of MSL's results exhibit both low mean and low variance, whereas TSL's results display both low mean and high variance, indicating a greater degree of uncertainty. This difference of uncertainty aligns with the insights derived from the route sheet as an intermediate workspace. These results suggest that our constraint-centric architecture delivers both exceptional and consistent performance across various scenarios, thereby meeting the requirements of the manufacturing community. The findings further reveal the potential of our architecture to democratize the automatic, entire-workflow constraint specification for production planning and scheduling for all manufacturing practitioners, ranging from \acp{oem} to \acp{sme}.

\section{Conclusion}\label{sec:con}

This paper addresses the critical challenge of automating constraint specification from heterogeneous raw data in smart manufacturing by introducing a constraint-centric architecture that effectively regulates \acp{llm} through domain-specific representations. Our three-level hierarchical structure successfully balances the generative capabilities of \acp{llm} with manufacturing reliability requirements. The automated adaptation algorithm enables efficient customization across different production scenarios, making the architecture broadly applicable for various manufacturing practices. Experimental results demonstrate the architecture's superiority over pure \ac{llm}-based approaches in maintaining precision while reducing reliance on human expert intervention. This work advances the practical implementation of smart manufacturing by providing a robust framework for automated constraint specification that meets the demands of modern production environments. Future research could explore the architecture's adaptation boundary to more complex manufacturing scenarios and its integration with other smart manufacturing systems.

\section*{Acknowledgment}

Q. Xu is a visiting student at Peking University from University of Science and Technology of China. The authors would like to thank Jiawen Liu for the assistance in figure drawing and the reviewers for insightful suggestions.

\bibliographystyle{ieeetr}
\bibliography{references}

\input{bios/bio.sty}

\end{document}